\documentclass[runningheads]{llncs}

\usepackage{caption}
\usepackage{xcolor}
\usepackage{booktabs}
\usepackage{url}
\usepackage[T1]{fontenc}
\usepackage{lmodern}
\usepackage{amsmath, amssymb}
\usepackage{graphicx}
\usepackage{array}
\usepackage{float}
\usepackage{multirow}
\usepackage{subcaption}
\usepackage{float}
\usepackage{tikz}
\usetikzlibrary{arrows.meta}
\usetikzlibrary{backgrounds} 
\pgfdeclarelayer{background}
\pgfsetlayers{background,main}
\usepackage{graphicx}
\usepackage{pgf}
\usepackage{lmodern}
\usepackage{import}
\usepackage{pgfplots}

\everymath=\expandafter{\the\everymath\displaystyle}
\IfFileExists{scrextend.sty}{
  \usepackage[fontsize=10.000000pt]{scrextend}
}{
  \renewcommand{\normalsize}{\fontsize{10.000000}{12.000000}\selectfont}
  \normalsize
}

\usetikzlibrary{fit}

\usepackage{mathalpha} 
\usepackage{amsmath}   
\usepackage{threeparttable}

\pgfplotsset{compat=1.18}

\begin{document}

\newcommand{\Loss}{\mathcal{L}}
\newcommand{\GateNotation}{\mathcal{G}}
\newcommand{\PoolFunc}{\mathcal{F}}

\title{\texttt{N-BEATS-MOE}: \texttt{N-BEATS} with a Mixture-of-Experts Layer for Heterogeneous Time~Series~Forecasting}

\titlerunning{\texttt{N-BEATS} with a Mixture-of-Experts Layer}
 
\author{Ricardo~Matos\inst{1}, Luis~Roque\inst{1,2}, 
Vitor~Cerqueira\inst{1,2}}

\authorrunning{Matos et al.}

\institute{
Faculdade de Engenharia da Universidade do Porto, Porto, Portugal \and
Laboratory for Artificial Intelligence and Computer Science (LIACC), Portugal\\ 
\email{ricardo.andre.de.matos@gmail.com}
\
}

\maketitle

\begin{abstract}

Deep learning approaches are increasingly relevant for time series forecasting tasks.  Methods such as \texttt{N-BEATS}, which is built on stacks of multilayer perceptrons (MLPs) blocks, have achieved state-of-the-art results on benchmark datasets and competitions. \texttt{N-BEATS} is also more interpretable relative to other deep learning approaches, as it decomposes forecasts into different time series components, such as trend and seasonality. In this work, we present \texttt{N-BEATS-MOE}, an extension of \texttt{N-BEATS} based on a Mixture-of-Experts (MoE) layer. \texttt{N-BEATS-MOE} employs a dynamic block weighting strategy based on a gating network which allows the model to better adapt to the characteristics of each time series. We also hypothesize that the gating mechanism provides additional interpretability by identifying which expert is most relevant for each series. We evaluate our method across 12 benchmark datasets against several approaches, achieving consistent improvements on several datasets, especially those composed of heterogeneous time series.

\keywords{Time series \and Forecasting \and Mixture-of-Experts \and \texttt{N-BEATS}}
\end{abstract}

\section{Introduction}

Time series forecasting plays a crucial role in numerous real-world applications. While traditional statistical methods such ARIMA~\cite{hyndman_forecasting_2018} have long been the standard approach, deep neural networks are increasingly showing their effectiveness in benchmark datasets and competitions~\cite{oreshkin_n-beats_2020}.

Deep learning approaches have shown particular success in forecasting problems involving multiple time series~\cite{challu_n-hits_2022,oreshkin_n-beats_2020}. These models can effectively learn patterns not only across time but also across collections of time series~\cite{montero-manso_principles_2021}. In real-world forecasting scenarios, datasets often contain heterogeneous time series with varying characteristics such as distinct trend or seasonal patterns or varying noise and scale levels. This issue motivates research into new neural architectures that can effectively handle such diversity. Additionally, having interpretable models that can explain their predictions is also a desirable property.

A notable state-of-the-art architecture in deep learning-based time series forecasting is \texttt{N-BEATS} (Neural Basis Expansion Analysis for Time Series)~\cite{oreshkin_n-beats_2020}. In its interpretable configuration, it is composed of a trend and seasonal stacks that are responsible for projecting the time series into a basis function modeling those components, allowing it to produce forecasts that are decomposable and, thus, more interpretable. \texttt{N-BEATS} has been effectively applied to datasets involving multiple time series~\cite{oreshkin_n-beats_2020,makridakis2022m5}.

In this work, we propose augmenting the \texttt{N-BEATS} architecture with a Mixture-of-Experts (MoE) layer\footnote{
\url{https://github.com/zaai-ai/mixture_of_experts_time_series}}. Our key modification is replacing the standard sum aggregation of block outputs with a weighted sum, where a gating network dynamically determines the weights. This allows the model to adaptively focus on different components depending on the input series, enhancing its ability to handle heterogeneous datasets with varying time series characteristics, such as different trend patterns and seasonal profiles.


Our working hypothesis is that the integration of an MoE layer not only enhances the \texttt{N-BEATS}'s performance by better handling heterogeneous time series but also improves transparency and explainability. When processing an input time series, the gating mechanism produces a set of weights that indicate which experts are most relevant for that particular series. These routing weights provide insights into which experts specialize in specific patterns or characteristics. Overall, this routing mechanism provides an additional layer of interpretability to \texttt{N-BEATS}.

We evaluate the proposed approach, and some variations, by conducting experiments on 12 benchmarks datasets comprising a total of 100.141 time series. The results indicate that \texttt{N-BEATS-MOE} shows a competitive performance with \texttt{N-BEATS}, especially in datasets composed of heterogeneous time series such as M3~\cite{makridakis_m3-competition_2000}. Moreover, a decomposition-based analysis of the results provided insights into the behaviour of the gating mechanism. 


\section{Background}\label{sec:background}

This section provides a background to our work. We start by introducing time series forecasting fundamentals in Section~\ref{sec:ts-forecasting}. We then discuss deep learning approaches for forecasting, with a focus on the \texttt{N-BEATS} architecture in Section~\ref{sec:dl-forecasting}. Finally, we review mixture-of-experts models and their applications in neural networks in Section~\ref{sec:moe}.

\subsection{Time Series Forecasting}\label{sec:ts-forecasting}

A univariate time series is defined as an ordered sequence of observations \(\{y_t\}_{t=1}^T\), where \(y_t\) represents the value of a single variable at time step \(t\), and \(T\) denotes the length of the series. The objective of time series forecasting is to predict future values \(\{y_{T+1}, y_{T+2}, \ldots, y_{T+H}\}\), where \(H\) is the forecast horizon, based on the historical observations.

Time series datasets often contain multiple individual series for forecasting. For example, in retail, organizations need to forecast sales for thousands of different products, each representing a different time series. Rather than building separate models for each series, global forecasting approaches aim to build a single model that can learn patterns across all time series in the dataset~\cite{januschowski_criteria_2020}. This allows the model to leverage information from the entire dataset to improve predictions for individual series.

\subsection{Deep Learning for forecasting}\label{sec:dl-forecasting}

Deep learning approaches are effective in forecasting problems, as evidenced by the state-of-the-art results on benchmark datasets and competitions such as M4~\cite{makridakis_m4_2018} and M5~\cite{makridakis2022m5}. Neural networks address forecasting tasks via supervised learning typically following an auto-regressive modeling approach. In effect, future values are modeled using past lags as input explanatory variables~\cite{oreshkin_n-beats_2020}.

Historically, most of the literature has focused on MLPs~\cite{tang1991time} or recurrent approaches (e.g., LSTM~\cite{smyl2020hybrid}) for time series forecasting using neural networks. More recently, research in deep learning for forecasting has expanded to explore architectures beyond these, including convolutional neural networks~\cite{wu2022timesnet}, transformers~\cite{lim2021temporal}, and other approaches.


Despite numerous approaches, recent advances have demonstrated that purely feedforward, MLP-based architectures can achieve state-of-the-art forecasting accuracy~\cite{oreshkin_n-beats_2020,challu_n-hits_2022,das2023long}. A pioneering architecture in this direction is \texttt{N-BEATS}~\cite{oreshkin_n-beats_2020}.

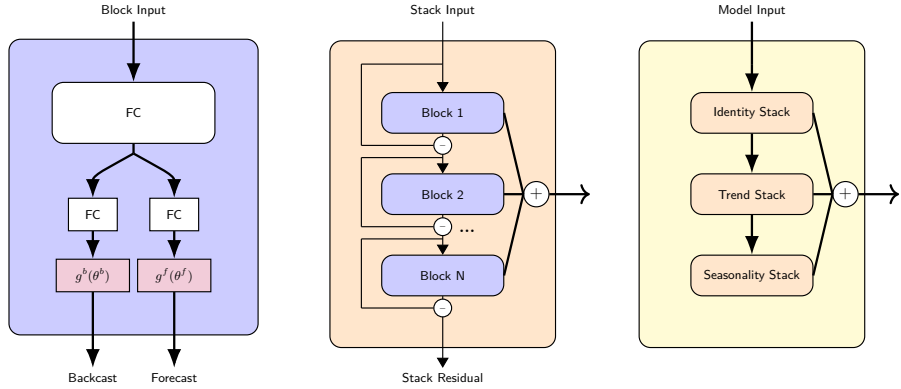
\begin{figure}[thb]
\resizebox{\textwidth}{!}{
\begin{tikzpicture}[font=\sffamily,
    arr/.style={-Latex, thick},
    arr_thin/.style={-Triangle, thin},
    container/.style={draw, fill=blue!20, rounded corners, inner sep=15pt},
    container_stack/.style={draw, fill=yellow!20, rounded corners, inner sep=18pt},
    container_residual/.style={draw, fill=orange!20, rounded corners, inner sep=18pt},
    mainblock/.style={draw, fill=white, rounded corners, minimum width=4cm, minimum height=1cm},
    mainstack/.style={draw, fill=orange!20, rounded corners, minimum width=3cm, minimum height=1cm},
    mainblock_2/.style={draw, fill=blue!20, rounded corners, minimum width=3cm, minimum height=1cm},
    gate/.style={draw, fill=white!20, rounded corners, minimum width=0.5cm, minimum height=0.5cm},
    expert/.style={draw, fill=white, rounded corners, minimum width=4cm, minimum height=1.5cm},
    fc/.style={draw, fill=white, rectangle, minimum width=1.2cm, minimum height=0.8cm},
    gblock/.style={draw, fill=purple!20, rectangle, minimum width=1.8cm, minimum height=0.8cm},
    identifier/.style={font=\LARGE\bfseries, text=black!70!black}
  ]

  \def\i{0}
  \begin{scope}[xshift=\i*3.8cm, scale=0.50, transform shape]]
      \node (in\i) at (0,4) {Block Input};

      \node[expert] (main\i) at (0,1.5) {FC};
      \draw[arr] (in\i.south) -- (main\i.north);

      \coordinate (split\i) at (0,0.5);
      \draw[thick] (main\i.south) -- (split\i);

      \node[fc] (fc1\i) at (-1,-1) {FC};
      \node[fc] (fc2\i) at (1,-1) {FC};
      \draw[arr] (split\i) to[out=-150,in=90] (fc1\i.north);
      \draw[arr] (split\i) to[out=-30,in=90] (fc2\i.north);

      \node[gblock] (gb\i) at (-1,-2.5) {$g^b(\theta^b)$};
      \node[gblock] (gf\i) at (1,-2.5) {$g^f(\theta^f)$};
      \draw[arr] (fc1\i.south) -- (gb\i.north);
      \draw[arr] (fc2\i.south) -- (gf\i.north);

      \node (backtxt\i) at (-1,-5) {Backcast};
      \node (fortxt\i) at (1,-5) {Forecast};
      \draw[arr] (gb\i.south) -- (backtxt\i.north);
      \draw[arr] (gf\i.south) -- (fortxt\i.north);


      \begin{pgfonlayer}{background}
        \node[container, fit=(main\i)(split\i)(fc1\i)(fc2\i)(gb\i)(gf\i)] {};
      \end{pgfonlayer}
    \end{scope}


    \def\i{1}
    \begin{scope}[xshift=\i*3.8cm, scale=0.50, transform shape]]
      \node (in\i) at (0,4) {Stack Input};
      \node[mainblock_2] (identity\i) at (0,1.5) {Block 1};

      \coordinate (split_1\i) at (0,2.7);
      \coordinate (split_2\i) at (-2,2.7);
      \coordinate (split_3\i) at (-2,0.7);
      
      \draw[thin] (in\i.south) -- (split_1\i.north);
      
      \draw[arr_thin] (split_1\i.south) -- (identity\i.north);
      \draw[thin] (split_1\i.west) -- (split_2\i.east);
      \draw[thin] (split_2\i.south) -- (split_3\i.north);

      \node[mainblock_2] (trend\i) at (0,-0.5) {Block 2};
      \draw[arr_thin] (identity\i.south) -- (trend\i.north);
      
      \node[mainblock_2] (seasonality\i) at (0,-2.5) {Block N};
      \draw[arr_thin] (trend\i.south) -- (seasonality\i.north);
      
      \node (backtxt\i) at (0,-5) {Stack Residual};
      \draw[arr_thin] (seasonality\i.south) -- (backtxt\i.north);
        
      \node[identifier, circle, draw=black, fill=white, minimum size=4pt, inner sep=2pt] (minus_1) at (0,0.7) {{\tiny$-$}};
      \draw[thin] (split_3\i.east) -- (minus_1.west);

      \coordinate (split_4\i) at (0,0.4);
      \coordinate (split_5\i) at (-2,0.4);
      \coordinate (split_6\i) at (-2,-1.3);

      \draw[thin] (split_4\i.west) -- (split_5\i.east);
      \draw[thin] (split_5\i.south) -- (split_6\i.north);

      \node[identifier, circle, draw=black, fill=white, minimum size=4pt, inner sep=2pt] (minus_2) at (0,-1.3) {{\tiny$-$}};
      \draw[thin] (split_6\i.east) -- (minus_2.west);

      \node[font=\bfseries\large, text=black] (text) at (0.6, -1.4) {...};

      \coordinate (split_7\i) at (0,-1.6);
      \coordinate (split_8\i) at (-2,-1.6);
      \coordinate (split_9\i) at (-2,-3.3);

      \draw[thin] (split_7\i.west) -- (split_8\i.east);
      \draw[thin] (split_8\i.south) -- (split_9\i.north);

      \node[identifier, circle, draw=black, fill=white, minimum size=4pt, inner sep=2pt] (minus_3) at (0,-3.3) {{\tiny$-$}};
      \draw[thin] (split_9\i.east) -- (minus_3.west);


      \node[identifier, circle, draw=black, fill=white, minimum size=14pt, inner sep=2pt] (split\i) at (2.3,-0.5) {\textbf{\large$+$}};
      
      \draw[-, thick, black] (identity\i.east) -- (split\i.west);
      \draw[-, thick, black] (trend\i.east) -- (split\i.west);
      \draw[-, thick, black] (seasonality\i.east) -- (split\i.west);

    \draw[->, thick, black] (split\i.east) -- ++(1,0) node[right] {};

      \begin{pgfonlayer}{background}
        \node[container_residual, fit=(identity\i)(trend\i)(seasonality\i)] {};
      \end{pgfonlayer}
    \end{scope}

  \def\i{2}
  \begin{scope}[xshift=\i*3.8cm, scale=0.50, transform shape]]
      \node (in\i) at (0,4) {Model Input};
      \node[mainstack] (identity\i) at (0,1.5) {Identity Stack};

      \coordinate (split_1\i) at (0,2.7);
      \draw[thick] (in\i.south) -- (split_1\i.north);
      \draw[arr] (split_1\i.south) -- (identity\i.north);

      \node[mainstack] (trend\i) at (0,-0.5) {Trend Stack};
      \draw[arr] (identity\i.south) -- (trend\i.north);
      
      \node[mainstack] (seasonality\i) at (0,-2.5) {Seasonality Stack};
      \draw[arr] (trend\i.south) -- (seasonality\i.north);


      \node[identifier, circle, draw=black, fill=white, minimum size=14pt, inner sep=2pt] (split\i) at (2.3,-0.5) {\textbf{\large$+$}};
      
      \draw[-, thick, black] (identity\i.east) -- (split\i.west);
      \draw[-, thick, black] (trend\i.east) -- (split\i.west);
      \draw[-, thick, black] (seasonality\i.east) -- (split\i.west);

    \draw[->, thick, black] (split\i.east) -- ++(1,0) node[right] {};

      \begin{pgfonlayer}{background}
        \node[container_stack, fit=(identity\i)(trend\i)(seasonality\i)] {};
      \end{pgfonlayer}
    \end{scope}

\end{tikzpicture}
}
    \caption{Architecture of the \texttt{N-BEATS} interpretable configuration originally proposed by Oreshkin et al.~\cite{oreshkin_n-beats_2020}. }

    \label{fig:n-beats-arch-pic}
\end{figure}

\texttt{N-BEATS} consists of multiple stacks of MLPs, each representing distinct basis functions, as illustrated in Figure~\ref{fig:n-beats-arch-pic}. Every stack comprises \(N\) blocks, each tasked with projecting the backward \(\boldsymbol{\theta}^b\) and forward \(\boldsymbol{\theta}^f\) expansion coefficients onto their respective basis functions. This process produces, for each block \(\ell\), the forecast and backcast outputs denoted by \(\hat{\mathbf{y}}_\ell = g_f^\ell(\boldsymbol{\theta}_\ell^f)\) and \(\hat{\mathbf{x}}_\ell = g_b^\ell(\boldsymbol{\theta}_\ell^b)\), respectively. Afterwards, the backcast \(\hat{\mathbf{x}}_\ell\) is subtracted from the block input signal \(\mathbf{x}_\ell\), and the resulting residual is passed to the subsequent block \(\ell + 1\). The forecast \(\hat{\mathbf{y}}_\ell\) is aggregated with previous forecasts as \(\hat{\mathbf{y}}_1 + \hat{\mathbf{y}}_2 + \cdots + \hat{\mathbf{y}}_\ell\).

In its interpretable configuration, \texttt{N-BEATS} is composed of a trend and seasonal stacks. In the trend stack, the model uses a polynomial basis \(\mathbf{B}(t)\), while in the seasonal stack it uses a harmonic Fourier basis \(\mathbf{F}(t)\), both applied over normalized time windows~\cite{oreshkin_n-beats_2020}:

\[
t = \frac{1}{H} \left[0, 1, 2, \ldots, H-2, H-1 \right],
\]
\[
\mathbf{B}_{i,:} = \begin{bmatrix}
1 & t_i & t_i^2 & \cdots & t_i^d
\end{bmatrix},
\]
\[
\mathbf{F}_{i,:} = \begin{bmatrix}
1 & \cos(2\pi t_i) & \cdots & \cos\left(2\pi \left\lfloor \frac{H}{2} + 1 \right\rfloor t_i \right), \sin(2\pi t_i) & \cdots & \sin\left(2\pi \left\lfloor \frac{H}{2} + 1 \right\rfloor t_i \right)
\end{bmatrix}
\]

Here, \(i \in \{0, \ldots, H{-}1\}\) indexes the time steps, and  \(t_i \in \{0, \frac{1}{H}, \ldots, \frac{H{-}1}{H}\}\)  represents normalized time. The parameter \( d \) is the degree of the polynomial in the trend basis.


\subsection{Mixture of experts}\label{sec:moe}

MoE methods address machine learning problems using an ensemble-based approach. The idea is to create multiple models, denoted as experts, and specialize them in different parts of the input space. The specialization process is governed by a gating neural network that learns to route inputs to the most appropriate experts~\cite{jacobs_adaptive_1991}.

A MoE typically consists of $N$ expert networks $\{f_1, \dots, f_N\}$ trained alongside a gating network $\GateNotation$. Originally proposed by Jacobs et al.~\cite{jacobs_adaptive_1991}, MoE aligns closely with the divide-and-conquer approach, a strategy that breaks down a complex problem into smaller, more manageable subproblems that are easier to solve. Unlike conventional models that activate all parameters for every input, MoEs dynamically select the active parameters depending on the input. This enables the model to better adapt to datasets that involve multiple domains, such as time series datasets composed of heterogeneous time series~\cite{makridakis_m4_2018}.


Although MoE approaches were first introduced decades ago~\cite{jacobs_adaptive_1991,masoudnia2014mixture}, they have gained popularity in the last few years, particularly in the context of LLMs (large language models)~\cite{dai_deepseekmoe_2024,jiang_mixtral_2024}.

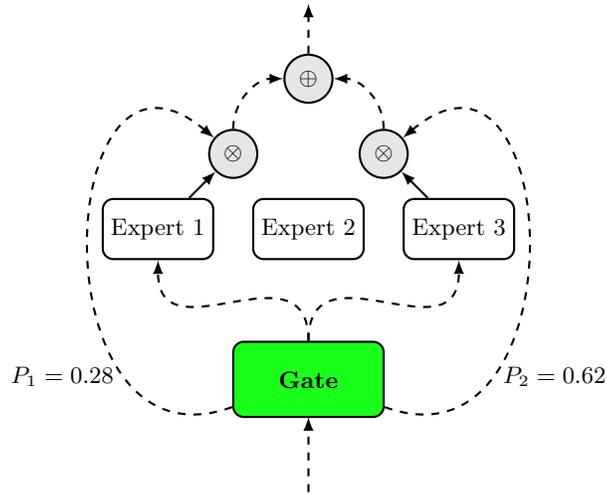
\begin{figure}[tb]
     \centering
    \begin{tikzpicture}[node distance=1.5cm and 1.5cm, >=latex, thick]
    
    \node[draw, rounded corners, fill=green!90, text=black, minimum width=2cm, minimum height=1cm] 
    (gate) at (0, 0) {\textbf{Gate}};
    
    \draw[->, thick, dashed] (0, -1.5) to[out=90, in=0, looseness=0] (0,-0.5);
    
    \node[draw, rectangle,  rounded corners, minimum width=1.2cm, minimum height=0.8cm, align=center] 
    (expert1) at (-2, 2) {Expert 1};
    \node[draw, rectangle,  rounded corners, minimum width=1.2cm, minimum height=0.8cm, align=center] 
    (expert2) at (0, 2) {Expert 2};
    \node[draw, rectangle,  rounded corners, minimum width=1.2cm, minimum height=0.8cm, align=center] 
    (expert3) at (2, 2) {Expert 3};
    
    \node[circle, draw, minimum size=0.6cm, fill=gray!20] 
    (mul1) at (-1, 3) {\small $\otimes$};
    \node[circle, draw, minimum size=0.6cm, fill=gray!20] 
    (mul2) at (1, 3) {\small $\otimes$};
    
    
    \draw[->, thick, dashed] (gate) to[out=90, in=-90, looseness=1.5] (expert1);
    \draw[->, thick, dashed] (gate) to[out=90, in=-90, looseness=1.5]  (expert3);
    
    \draw[->, thick] (expert1) -- (mul1);
    \draw[->, thick] (expert3) -- (mul2);
    
    \draw[->, thick, dashed] (gate) to[out=-160, in=140, looseness=2] node[near start, left, font=\small] {$P_1 = 0.28$}  (mul1);
    \draw[->, thick, dashed] (gate) to[out=-20, in=40, looseness=2] node[near start, right, font=\small] {$P_2 = 0.62$}  (mul2);
    
    \node[circle, draw, minimum size=0.6cm, fill=gray!20] 
    (sum) at (0, 4) {\small $\oplus$};
    
    \draw[->, thick, dashed] (mul1) to[out=90, in=180, looseness=1] (sum);
    \draw[->, thick, dashed] (mul2) to[out=90, in=0, looseness=1] (sum);
    
    \draw[->, thick, dashed] (sum) to[out=90, in=0, looseness=0] (0, 5);
    
    \end{tikzpicture}

    \caption{A sparse MoE consisting of 3 experts. The gating mechanism first selects the 2 most appropriate experts for the input based on their gate function scores. Then, a linear combination is performed, using the \textit{softmax} values of the gating scores as weights. The key distinction between sparse MoE and dense MoE is that, as shown in the image, not all experts are activated.} \label{fig:sparse-moe}
\end{figure}

MoEs are often applied in a sparse manner as illustrated in Figure~\ref{fig:sparse-moe}. That is, at any given point, only the \texttt{top\_k} best-suited experts for a given input are used, effectively reducing computational costs. Sparse approaches often come with a setback; it is common in this setting for the gating network to assign disproportionately high probabilities to a single expert, neglecting the remaining experts.  This imbalance can significantly restrict training opportunities for underutilized experts, further degrading the problem known as routing collapse~\cite{shazeer_outrageously_2017}.

Several approaches have been developed to solve this problem. Some studies~\cite{dai_deepseekmoe_2024,shi_time-moe_2025} introduce an auxiliary loss term designed to penalize uneven utilization of experts. In contrast, other approaches prioritize methods that avoid modifying the loss function, as such modifications often induce instability and result in suboptimal solutions.  For example, Zhou et al.~\cite{zhou_mixture--experts_2022} propose to reverse the traditional routing mechanism by allowing experts to select \textit{top-k} tokens (parts of the input in natural language processing tasks), rather than having a gate select the \textit{top-k} experts. More recently, Wang et al.~\cite{wang_auxiliary-loss-free_2024} introduce expert-wise biases applied directly to the routing scores which promotes balanced expert selection in the architecture rather than via loss penalties.

MoE approaches have been extensively explored in domains such as natural language processing~\cite{shazeer_outrageously_2017,du2022glam} or computer vision~\cite{v-moe,robust,MEGAN}. For time series forecasting, recent studies have shown that MoE is a promising approach. For example, two recent approaches, namely Time-MoE~\cite{shi_time-moe_2025} and Moirai~\cite{liu_moirai-moe_2024}, leverage MoE to build large scale time series foundation models. 
Other works have shown that MoE can be effective approaches in different forecasting scenarios or domains, such as long-term forecasting~\cite{ni_mixture--linear-experts_2024}, intelligent transportation systems~\cite{wang_time_2024}, multivariate time series~\cite{han2024kan4tsf}, distribution shifts~\cite{sun_learning_2024}, or pre-trained experts~\cite{cerqueira2019arbitrage}.

\section{Methodology}\label{sec:methodology}

In this section, we present our methodology for extending \texttt{N-BEATS}~\cite{oreshkin_n-beats_2020} with a MoE layer. \texttt{N-BEATS} is a state-of-the-art deep learning approach for time series forecasting that employs a modular architecture composed of stacks of MLPs. Our proposed approach, dubbed \texttt{N-BEATS-MOE}, aims to improve \texttt{N-BEATS} by introducing an adaptive weighting mechanism that learns to combine the outputs of these stacks.

\begin{figure}[hbt]
\centering
\begin{tikzpicture}[font=\sffamily,
    arr/.style={-Latex, thick},
    arr_thin/.style={-Triangle, thin},
    container/.style={draw, fill=blue!20, rounded corners, inner sep=15pt},
    container_stack/.style={draw, fill=yellow!20, rounded corners, inner sep=18pt},
    container_residual/.style={draw, fill=orange!20, rounded corners, inner sep=18pt},
    mainblock/.style={draw, fill=white, rounded corners, minimum width=4cm, minimum height=1cm},
    mainstack/.style={draw, fill=orange!20, rounded corners, minimum width=3cm, minimum height=1cm},
    mainblock_2/.style={draw, fill=blue!20, rounded corners, minimum width=3cm, minimum height=1cm},
    gate/.style={draw, fill=white!20, rounded corners, minimum width=0.5cm, minimum height=0.5cm},
    expert/.style={draw, fill=white, rounded corners, minimum width=4cm, minimum height=1.5cm},
    fc/.style={draw, fill=white, rectangle, minimum width=1.2cm, minimum height=0.8cm},
    gblock/.style={draw, fill=purple!20, rectangle, minimum width=1.8cm, minimum height=0.8cm},
    identifier/.style={font=\LARGE\bfseries, text=black!70!black}
  ]

      \def\i{0}
    \begin{scope}[xshift=\i*3.8cm, scale=0.50, transform shape]]
      \node (in\i) at (0,4) {Stack Input};
      \node[mainblock_2] (identity\i) at (0,1.5) {Block 1};

      \coordinate (split_1\i) at (0,2.7);
      \coordinate (split_2\i) at (-2,2.7);
      \coordinate (split_3\i) at (-2,0.7);
      
      \draw[thin] (in\i.south) -- (split_1\i.north);
      
      \draw[arr_thin] (split_1\i.south) -- (identity\i.north);
      \draw[thin] (split_1\i.west) -- (split_2\i.east);
      \draw[thin] (split_2\i.south) -- (split_3\i.north);

      \node[mainblock_2] (trend\i) at (0,-0.5) {Block 2};
      \draw[arr_thin] (identity\i.south) -- (trend\i.north);
      
      \node[mainblock_2] (seasonality\i) at (0,-2.5) {Block N};
      \draw[arr_thin] (trend\i.south) -- (seasonality\i.north);
      
      \node (backtxt\i) at (0,-5) {Stack Residual};
      \draw[arr_thin] (seasonality\i.south) -- (backtxt\i.north);
        
      \node[identifier, circle, draw=black, fill=white, minimum size=4pt, inner sep=2pt] (minus_1) at (0,0.7) {{\tiny$-$}};
      \draw[thin] (split_3\i.east) -- (minus_1.west);

      \coordinate (split_4\i) at (0,0.4);
      \coordinate (split_5\i) at (-2,0.4);
      \coordinate (split_6\i) at (-2,-1.3);

      \draw[thin] (split_4\i.west) -- (split_5\i.east);
      \draw[thin] (split_5\i.south) -- (split_6\i.north);

      \node[identifier, circle, draw=black, fill=white, minimum size=4pt, inner sep=2pt] (minus_2) at (0,-1.3) {{\tiny$-$}};
      \draw[thin] (split_6\i.east) -- (minus_2.west);

      \node[font=\bfseries\large, text=black] (text) at (0.6, -1.4) {...};

      \coordinate (split_7\i) at (0,-1.6);
      \coordinate (split_8\i) at (-2,-1.6);
      \coordinate (split_9\i) at (-2,-3.3);

      \draw[thin] (split_7\i.west) -- (split_8\i.east);
      \draw[thin] (split_8\i.south) -- (split_9\i.north);

      \node[identifier, circle, draw=black, fill=white, minimum size=4pt, inner sep=2pt] (minus_3) at (0,-3.3) {{\tiny$-$}};
      \draw[thin] (split_9\i.east) -- (minus_3.west);


      
      \draw[->, thick, black] (identity\i.east) -- ++(2,0);
      \draw[->, thick, black] (trend\i.east) -- ++(2,0);
      \draw[->, thick, black] (seasonality\i.east) -- ++(2,0);


      \begin{pgfonlayer}{background}
        \node[container_residual, fit=(identity\i)(trend\i)(seasonality\i)] {};
      \end{pgfonlayer}
    \end{scope}

  \def\i{1}
  \begin{scope}[xshift=\i*3.8cm, scale=0.50, transform shape]]
      \node (in\i) at (0,4) {Model Input};
      \node[mainstack] (identity\i) at (0,1.5) {Identity Stack};

      \coordinate (split_1\i) at (0,2.7);
      \draw[thick] (in\i.south) -- (split_1\i.north);
      \draw[arr] (split_1\i.south) -- (identity\i.north);

      \node[mainstack] (trend\i) at (0,-0.5) {Trend Stack};
      \draw[arr] (identity\i.south) -- (trend\i.north);
      
      \node[mainstack] (seasonality\i) at (0,-2.5) {Seasonality Stack};
      \draw[arr] (trend\i.south) -- (seasonality\i.north);

      \node[gate, identifier] (gate\i) at (2.3, 2.7) {\(\GateNotation\)};


      \node[identifier, circle, draw=black, fill=white, minimum size=14pt, inner sep=2pt] (split\i) at (2.3,-0.5) {\textbf{\large$\sum$}};
      
      \draw[-, thick, black] (identity\i.east) -- (split\i.west);
      \draw[-, thick, black] (trend\i.east) -- (split\i.west);
      \draw[-, thick, black] (seasonality\i.east) -- (split\i.west);

    \draw[dashed, -] (gate\i.south) -- (split\i.north);

    \draw[->, thick, black] (split_1\i.east) -- (gate\i.west);
    
    \draw[->, thick, black] (split\i.east) -- ++(1,0) node[right] {};

      \begin{pgfonlayer}{background}
        \node[container_stack, fit=(identity\i)(trend\i)(seasonality\i)] {};
      \end{pgfonlayer}
    \end{scope}

\end{tikzpicture}
    \caption{The proposed architecture,  \texttt{N-BEATS-MOE}. The gate $\GateNotation$ is responsible for determining the importance of each block based on the input data.}
    \label{fig:arch-pic}
\end{figure}
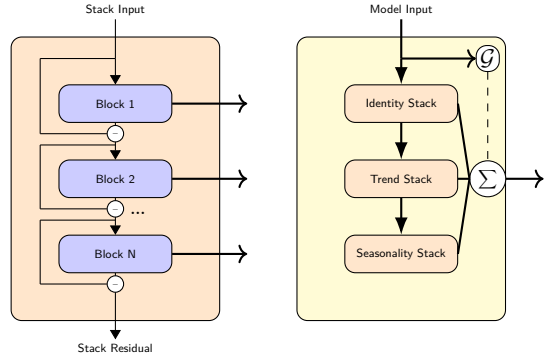

In our proposed method (Figure~\ref{fig:arch-pic}), we replace the standard summation of block outputs in \texttt{N-BEATS} with a MoE layer. In this setup, each block acts as an expert, and the gating mechanism learns to determine the appropriate contribution or importance of each to the final output.

Formally, let the outputs of each of the \( B \) blocks in the model be denoted by \(\hat{\mathbf{y}}_{\ell}\), where \( \ell \in \{1, \dots, B\} \), corresponding to forecast contributions from stacks such as \textit{identity}, \textit{trend}, and \textit{seasonal}. In the original \texttt{N-BEATS} architecture, these outputs are simply summed to produce the final forecast:

\begin{equation}
    \hat{\mathbf{y}} = \sum_{\ell=1}^{B} \hat{\mathbf{y}}_{\ell}
\end{equation}

In our proposed \texttt{N-BEATS-MOE} approach, this summation is replaced by a weighted combination using a learned gating mechanism, yielding:

\begin{equation}
    \hat{\mathbf{y}} = \sum_{\ell=1}^{B} \hat{\GateNotation}_{\ell} \cdot \hat{\mathbf{y}}_{\ell},
\end{equation}

    where the block gating weights \(\hat{\GateNotation}_{\ell}\) are defined as:

\begin{equation}
\hat{\GateNotation}_{\ell} = 
\mathrm{softmax}_\ell\left( \mathrm{LINEAR}_\ell(\mathbf{x}_{0}) \right).
\end{equation}

Here, \(\mathrm{LINEAR}_\ell\) denotes a learnable affine transformation applied to the model input \(\mathbf{x}_0\), which is first normalized through a LayerNorm to prevent mode collapse. The gating weights assign different importance to each block output, allowing the model to adaptively combine them.

This dynamic weighting mechanism enables the model to prioritize stack types based on the characteristics of the input series aiming at improving performance on heterogeneous datasets with diverse temporal patterns. Besides a better adaptability, this approach also allows for a better interpretability and explainability, since the output of the gating network can provide insights about which stack types are most important for a given input.

\section{Experiments}

This section describes the experiments conducted to validate \texttt{N-BEATS-MOE}. These are designed to address the following research questions:

\begin{itemize}
    \item \textbf{RQ1}: How can we augment the \texttt{N-BEATS} architecture with a MoE approach to better accommodate the \textit{heterogeneous patterns} (e.g., varying trends and seasonal profiles) inherent in datasets comprising multiple univariate time series? This question is addressed with the proposed methodology presented in Section~\ref{sec:methodology}.

    \item \textbf{RQ2}: To what extent do \texttt{N-BEATS} MoE-augmented approaches achieve better performance in terms of forecast accuracy? 
   
    \item \textbf{RQ3}: How does the inclusion of a MoE layer improve the interpretability of \texttt{N-BEATS}?

\end{itemize}

\subsection{Dataset and Evaluation}

We evaluate forecasting performance using 12 time series datasets originating from 4 forecasting competitions:  M1~\cite{makridakis_accuracy_1982}, Tourism~\cite{athanasopoulos_tourism_2011}, M3~\cite{makridakis_m3-competition_2000}, and M4~\cite{makridakis_m4_2018}. These cover three frequencies: monthly, quarterly, and yearly. Three of these four sources contain time series from varied application domains, which provide an adequate test bed for evaluating the performance of the proposed approach in handling heterogeneous time series. Table~\ref{tab:datasets_summary} summarizes the datasets.

\begin{table}[ht]
\centering
\small
\setlength{\tabcolsep}{6pt} 
\renewcommand{\arraystretch}{0.9} 
\caption{Summary of datasets by frequency}
\label{tab:datasets_summary}
\begin{tabular}{lllr}
\hline
\textbf{Dataset} & \textbf{Frequency} & \textbf{Count} & \textbf{Domains} \\
\hline
\multirow{3}{*}{M1}      
         & Yearly    & 181   & \multirow{3}{*}{Mixed}    \\
         & Quarterly & 203   &                          \\
         & Monthly   & 617   &                          \\
\hline
\multirow{3}{*}{Tourism} 
         & Yearly    & 518   &  \multirow{3}{*}{Tourism}                         \\
         & Quarterly & 427   &                          \\
         & Monthly   & 366   &                          \\
\hline
\multirow{3}{*}{M3}      
         & Yearly    & 645   &  \multirow{3}{*}{Mixed}                         \\
         & Quarterly & 756   &                          \\
         & Monthly   & 1428  &                          \\
\hline
\multirow{3}{*}{M4}      
         & Yearly    & 23000 &  \multirow{3}{*}{Mixed}                         \\
         & Quarterly & 24000 &                          \\
         & Monthly   & 48000 &                          \\
\hline
\end{tabular}
\end{table}

We use SMAPE (Symmetric Mean Absolute Percentage Error) as the evaluation metric, which is defined as follows:

\begin{equation}
\frac{1}{H} \sum_{\tau=t+1}^{t+H} \frac{\left| y_{\tau} - \hat{y}_{\tau} \right|}{\left| y_{\tau} \right| + \left| \hat{y}_{\tau} \right|}
\end{equation}

\noindent where \( H \) represents the forecasting horizon, \( y_{\tau} \) is the actual value at time \( \tau \), and \( \hat{y}_{\tau} \) is the predicted value. 

\subsection{Methods}

\begin{figure}[htb]
\resizebox{\textwidth}{!}{
\begin{tikzpicture}[font=\sffamily,
    arr/.style={-Latex, thick},
    container/.style={draw, fill=blue!20, rounded corners, inner sep=15pt},
    container_stack/.style={draw, fill=yellow!20, rounded corners, inner sep=18pt},
    mainblock/.style={draw, fill=white, rounded corners, minimum width=4cm, minimum height=1cm},
    mainstack/.style={draw, fill=orange!20, rounded corners, minimum width=3cm, minimum height=1cm},
    gate/.style={draw, fill=white!20, rounded corners, minimum width=0.5cm, minimum height=0.5cm},
    expert/.style={draw, fill=white, rounded corners, minimum width=4cm, minimum height=1.5cm},
    fc/.style={draw, fill=white, rectangle, minimum width=1.2cm, minimum height=0.8cm},
    gblock/.style={draw, fill=purple!20, rectangle, minimum width=1.8cm, minimum height=0.8cm},
    identifier/.style={font=\LARGE\bfseries, text=black!70!black}
  ]

  \def\i{0}
  \begin{scope}[xshift=\i*3.8cm, scale=0.50, transform shape]]
      \node (in\i) at (0,4) {Block Input};
      \node[expert] (main\i) at (0.4,1.8) {Expert 3};
      \node[expert] (main\i) at (0.2,1.65) {Expert 2};
      \node[expert] (main\i) at (0,1.5) {Expert 1};
      \draw[arr] (in\i.south) -- (main\i.north);

      \coordinate (split\i) at (0,0.5);
      \draw[thick] (main\i.south) -- (split\i);

      \node[fc] (fc1\i) at (-1,-1) {\(\theta_b\)};
      \node[fc] (fc2\i) at (1,-1) {\(\theta_f\)};
      \draw[arr] (split\i) to[out=-150,in=90] (fc1\i.north);
      \draw[arr] (split\i) to[out=-30,in=90] (fc2\i.north);

      \node[gblock] (gb\i) at (-1,-2.5) {$g^b(\theta^b)$};
      \node[gblock] (gf\i) at (1,-2.5) {$g^f(\theta^f)$};
      \draw[arr] (fc1\i.south) -- (gb\i.north);
      \draw[arr] (fc2\i.south) -- (gf\i.north);

      \node (backtxt\i) at (-1,-5) {Backcast};
      \node (fortxt\i) at (1,-5) {Forecast};
      \draw[arr] (gb\i.south) -- (backtxt\i.north);
      \draw[arr] (gf\i.south) -- (fortxt\i.north);

      \node[identifier] (identifier\i) at (0, -6) {2) MoEBlock};

      \begin{pgfonlayer}{background}
        \node[container, fit=(main\i)(split\i)(fc1\i)(fc2\i)(gb\i)(gf\i)] {};
      \end{pgfonlayer}
    \end{scope}

  \def\i{1}
  \begin{scope}[xshift=\i*3.8cm, scale=0.50, transform shape]]
      \node (in\i) at (0,4) {Block Input};

      \coordinate (split_shared\i) at (0,2.8);
      \draw[thick] (in\i.south) -- (split_shared\i);
      
      \begin{scope}[xshift=1.4cm,scale=0.6, transform shape]]
          \node[expert] (main_e\i) at (0.4,2.8) {Expert 3};
          \node[expert] (main_e\i) at (0.2,2.65) {Expert 2};
          \node[expert] (main_e\i) at (0,2.5) {Expert 1};
      \end{scope}
      \begin{scope}[xshift=-1.2cm,scale=0.6, transform shape]]
          \node[expert] (main_s\i) at (0,2.5) {Shared Expert};
      \end{scope}
      \draw[arr] (split_shared\i) -- (main_e\i.north);
      \draw[arr] (split_shared\i) -- (main_s\i.north);

      \node[circle, draw=black, fill=white, minimum size=12pt, inner sep=0pt] (split\i) at (0,0.5) {+};

      \draw[thick] (main_s\i.south) -- (split\i);
      \draw[thick] (main_e\i.south) -- (split\i);

      \node[fc] (fc1\i) at (-1,-1) {\(\theta_b\)};
      \node[fc] (fc2\i) at (1,-1) {\(\theta_f\)};
      \draw[arr] (split\i) to[out=-150,in=90] (fc1\i.north);
      \draw[arr] (split\i) to[out=-30,in=90] (fc2\i.north);

      \node[gblock] (gb\i) at (-1,-2.5) {$g^b(\theta^b)$};
      \node[gblock] (gf\i) at (1,-2.5) {$g^f(\theta^f)$};
      \draw[arr] (fc1\i.south) -- (gb\i.north);
      \draw[arr] (fc2\i.south) -- (gf\i.north);

      \node (backtxt\i) at (-1,-5) {Backcast};
      \node (fortxt\i) at (1,-5) {Forecast};
      \draw[arr] (gb\i.south) -- (backtxt\i.north);
      \draw[arr] (gf\i.south) -- (fortxt\i.north);

      \node[identifier] (identifier\i) at (0, -6) {3) MoEShared};

      \begin{pgfonlayer}{background}
        \node[container, fit=(main_e\i)(main_s\i)(split\i)(fc1\i)(fc2\i)(gb\i)(gf\i)] {};
      \end{pgfonlayer}
    \end{scope}

  \def\i{2}
  \begin{scope}[xshift=\i*3.8cm, scale=0.50, transform shape]]
      \node (in\i) at (0,4) {Block Input};
      \node[draw, fill=white, rounded corners, minimum width=4.5cm, minimum height=2.2cm] (main3\i) at (0,1.5) {Expert 3};
    \node[draw, fill=white, rounded corners, minimum width=3.8cm, minimum height=1.8cm] (main2\i) at (0,1.5) {Expert 2};
    \node[draw, fill=white, rounded corners, minimum width=3.2cm, minimum height=1.5cm] (main1\i) at (0,1.5) {Expert 1};

      \draw[arr] (in\i.south) -- (main3\i.north);

      \coordinate (split\i) at (0,0.4);
      \draw[thick] (main3\i.south) -- (split\i);

      \node[fc] (fc1\i) at (-1,-1) {\(\theta_b\)};
      \node[fc] (fc2\i) at (1,-1) {\(\theta_f\)};
      \draw[arr] (split\i) to[out=-150,in=90] (fc1\i.north);
      \draw[arr] (split\i) to[out=-30,in=90] (fc2\i.north);

      \node[gblock] (gb\i) at (-1,-2.5) {$g^b(\theta^b)$};
      \node[gblock] (gf\i) at (1,-2.5) {$g^f(\theta^f)$};
      \draw[arr] (fc1\i.south) -- (gb\i.north);
      \draw[arr] (fc2\i.south) -- (gf\i.north);

      \node (backtxt\i) at (-1,-5) {Backcast};
      \node (fortxt\i) at (1,-5) {Forecast};
      \draw[arr] (gb\i.south) -- (backtxt\i.north);
      \draw[arr] (gf\i.south) -- (fortxt\i.north);

      \node[identifier] (identifier\i) at (0, -6) {4) MoEScaled};

      \begin{pgfonlayer}{background}
        \node[container, fit=(main1\i)(split\i)(fc1\i)(fc2\i)(gb\i)(gf\i)] {};
      \end{pgfonlayer}
    \end{scope}
    
\end{tikzpicture}
}
    \caption{\texttt{N-BEATS-MOE} variations. In these variations, we focus on applying the MoE at the block level. For clarity, the gating mechanism—a simple linear layer—is omitted.}
    \label{fig:arch-pic-variations}
\end{figure}
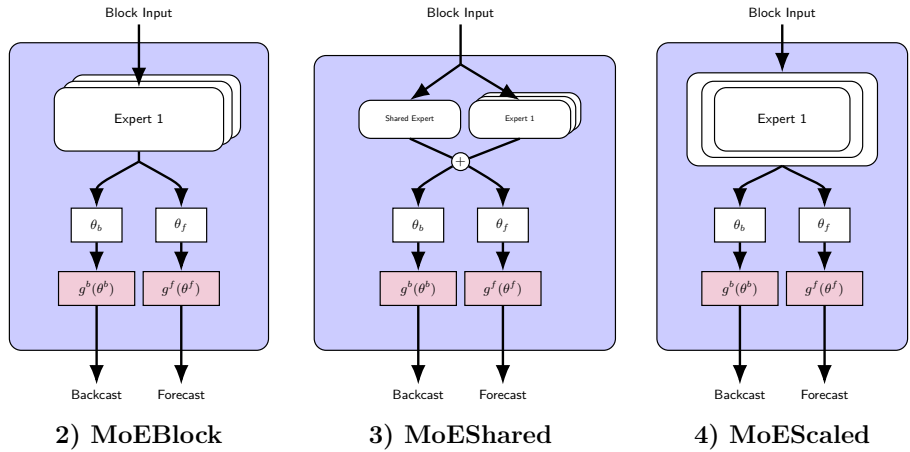

We compare \texttt{N-BEATS-MOE} with \texttt{N-BEATS}, seasonal naive, and several \texttt{N-BEATS-MOE} variants that add a MoE layer to \texttt{N-BEATS} using different strategies, as illustrated in Figure~\ref{fig:arch-pic-variations}:

\begin{itemize}
    \item \texttt{N-BEATS}-(MOEBlock): In this variation, we replace the fully-connected (FC) layer in the \texttt{N-BEATS} basic building blocks with a MoE layer consisting of \textit{N} expert FC layers, where each expert models different basis coefficients corresponding to subsets of series patterns. 

    \item \texttt{N-BEATS}-(MOEShared): Inspired by DeepSeek's approach~\cite{dai_deepseekmoe_2024}, this variant uses one shared expert for all inputs, together with specialized experts that the gating network activates based on the input data.

    \item \texttt{N-BEATS}-(MOEScaled): This variant uses experts with different parameter sizes to test whether model specialization based on time series complexity leads to improved performance.
\end{itemize}

\subsection{Training procedure}

Each dataset was divided into training, validation, and test subsets. The test set is composed of the last $H$ observations of each time series in a given dataset. The validation set is partitioned in the same way, being composed of the last $H$ observations of each time series before the test set.
Models were trained using the MAE loss function and the Adam optimizer with default parameters and a learning rate of 0.001. A StepLR scheduler with a gamma value of 0.5 was used to adjust the learning rate during training. Early stopping based on validation performance was used to prevent overfitting.


Hyperparameter tuning was performed using the training and validation subsets. A Bayesian optimization algorithm was used, with the number of trials set to 20. The hyperparameters that achieved the best performance in the validation set were then used to evaluate the model on the test set. Table~\ref{tab:hyper_table} summarizes the hyperparameter search space.

\begin{table}[t]
\centering
\begin{threeparttable}
\caption[Hyperparameter search space configuration used in experiments]{Hyperparameter search space configuration used in experiments.}
\label{tab:hyper_table}
\begin{tabular}{@{}lp{8cm}@{}}
\toprule
Parameter & Search Space Values \\
\midrule
\multicolumn{2}{@{}l}{Model Structure} \\
\midrule
\texttt{Input window multiplier} & 1, 2, 3, 4, 5 \\
\texttt{Forecast horizon (h)} & Dataset-dependent \\
\texttt{Stack types} & [\textit{``identity''}, \textit{``trend''}, \textit{``seasonality''}] \\
\texttt{MLP units} & $[2^i, 2^i]$ for $i \in \{2, 3, \ldots, 9\}$\tnote{a} \\
\texttt{Number of blocks} & $[m, m, m]$ for $m \in \{1, 3, 6, 9\}$ \\
\texttt{Scaler type} & \textit{``identity''} \\
\texttt{Shared weights} & True \\
\texttt{Number of experts} & $2^j$ for $j \in \{1, 2, 3\}$\tnote{b} \\
\texttt{Top-k experts} & $2^l$ for $l \in \{0, 1, 2, 3\}$\tnote{b} \\
\midrule
\multicolumn{2}{@{}l}{Training Parameters} \\
\midrule
\texttt{Max training steps} & 1000, 2500, 5000, 10000 \\
\texttt{Batch size} & 32, 64, 128, 256 \\
\texttt{Windows batch size} & 128, 256, 512, 1024 \\
\texttt{Early stopping patience } & 10, 20 \\
\bottomrule
\end{tabular}
\begin{tablenotes}
\small
\item[a] For \texttt{N-BEATS} models, $i \in \{2, 3, \ldots, 10\}$.
\item[b] Not applicable for \texttt{N-BEATS} and \texttt{N-BEATS-MOE} (set to 0).
\end{tablenotes}
\end{threeparttable}
\end{table}

\subsection{Results}

Table~\ref{tab:combined_smape} summarizes the SMAPE scores across all datasets and frequency groups, comparing \texttt{N-BEATS-MOE} with other variants, seasonal naive, and also the original \texttt{N-BEATS} method. We report median results over 10 trials to account for variability due to random initialization and data sampling.


On the M1 dataset, \texttt{N-BEATS-MOE} performs significantly better on yearly (9.76\%) and monthly (14.84\%) frequencies, with a clear improvement over \texttt{N-BEATS}. Most of other MoE-based variations also perform better than \texttt{N-BEATS}. On the M3 dataset, the results are more comparable and neither model is able to clearly stand out, with \texttt{N-BEATS} being the best on the yearly (16.3\%) and quarterly (9.01\%) frequencies, and \texttt{N-BEATS-MOE} the best on the monthly (13.96\%). For M4, \texttt{N-BEATS-MOE} is clearly the best in both yearly (13.31\%) and quarterly (9.82\%) frequencies but fails to perform better than \texttt{N-BEATS} on the monthly frequency. For the Tourism datasets, all MoE-based models perform worse than \texttt{N-BEATS}. 



\begin{table}[htb]
\centering
\caption[Experiment results]{SMAPE across datasets, models, and frequency groups. Values in \textbf{bold} (\textcolor{red}{red}) represent the best (second best) approach.}
\label{tab:combined_smape}
\resizebox{\textwidth}{!}{
\begin{tabular}{llcccccc}
\toprule
\textbf{Dataset} & \textbf{Freq./H.} & \textbf{\texttt{N-BEATS}} & \textbf{\texttt{N-BEATS-MOE}} & \textbf{MoEBlock} & \textbf{MoEShared} & \textbf{MoEScaled} & \textbf{Seas.Naive} \\
\midrule
\multirow{3}{*}{M1} 
    & Yearly/2     & 10.87 & \textbf{9.76} & 10.90 & \textcolor{red}{9.97} & 11.62 & 11.89\\
    & Quarterly/2  & 11.92 & \textcolor{red}{11.36} & 11.60 & \textbf{11.28} & 11.44 & 16.54\\
    & Monthly/8    & 15.53 & \textbf{14.84} & \textcolor{red}{15.17} & 15.57 & 15.49 & 16.57\\
\midrule
\multirow{3}{*}{Tourism} 
    & Yearly/4     & \textbf{26.48} & \textcolor{red}{26.51} & 27.46 & 26.77 & 29.78 & 27.66\\
    & Quarterly/8  & \textcolor{red}{19.29} & 20.96 & \textbf{19.28} & 19.95 & 19.45 & 21.10\\
    & Monthly/18    & \textbf{24.39} & 25.13 & 24.77 & \textcolor{red}{24.64} & 24.76 & 27.30\\
\midrule
\multirow{3}{*}{M3} 
    & Yearly/6     & \textbf{16.13} & 16.27 & 16.29 & \textcolor{red}{16.17} & 16.28 &  17.87\\
    & Quarterly/8  & \textbf{9.01} & \textcolor{red}{9.03} & 9.12 & 9.68 & 9.38 & 11.07\\
    & Monthly/18    & 14.03 & \textbf{13.96} & 14.08 & 14.07 & \textcolor{red}{13.99} & 17.24\\
\midrule
\multirow{3}{*}{M4} 
    & Yearly/6     & 13.45 & \textbf{13.31} & 13.48 & 13.36 & \textcolor{red}{13.34} & 16.34\\
    & Quarterly/8  & 9.92  & \textbf{9.82} & \textcolor{red}{9.87}  & 10.02 & 9.93 & 12.52\\
    & Monthly/18    & \textbf{12.79} & 12.99 & 13.10 & 13.12 & \textcolor{red}{12.95} & 15.99\\
\bottomrule
\end{tabular}
}
\end{table}

Overall, the results suggest that our approach provides consistent improvements over \texttt{N-BEATS}, especially in datasets composed of heterogeneous time series (i.e. time series from various domains). In these (M1, M3, and M4), \texttt{N-BEATS-MOE} outperforms \texttt{N-BEATS} in 6 out of 9 cases. However, in the Turism datasets, composed of time series from a single domain, the proposed approach loses in all three frequency variants.
Besides these results, \texttt{N-BEATS-MOE} outperforms seasonal naive in all cases, validating its forecasting accuracy.
Except for a few cases, the proposed method also shows better performance than its variants that employ a MoE layer differently.

\subsection{Expert Specialization Analysis}

\subsubsection{Analysis via STL Decomposition.}

To analyze expert specialization within the \texttt{N-BEATS-MOE} architecture, we conducted a controlled experiment using STL~\cite{cleveland1990stl} to decompose time series data, and evaluated the gating behavior across three expert stacks: trend, seasonal, and identity. The results are presented in Table~\ref{tab:expert-stats-full}, which shows the ratio of times each expert stack is assigned the highest probability by the gating network, split by component.

Using datasets M1, M3, and M4 (monthly frequency), we observed that in M3, the gating mechanism aligned well with the decomposition, accurately assigning trend and seasonal components to their respective experts. In M4, this alignment persisted for trend components but was weaker for seasonal ones, with the trend expert still often favored. Interestingly, in M1, the expert assignments diverged significantly from the expectations of the decomposition, though \texttt{N-BEATS-MOE} still achieved a superior overall performance over \texttt{N-BEATS}. These results suggest that while \texttt{N-BEATS-MOE} can learn to mirror structural decompositions, its gating behavior is also sensitive to dataset-specific characteristics and does not always align precisely with the expected decomposition patterns. 

\begin{table}[ht]
\centering 
\caption{Expert selection ratios by dataset and decomposition component for the monthly frequency. Values show the proportion of times each MoE expert was selected as most relevant for a given input.}
\label{tab:expert-stats-full}
\begin{tabular}{llccc}
\toprule
\textbf{Dataset} & \textbf{Component} & \textbf{Identity} & \textbf{Trend} & \textbf{Seasonal} \\
\midrule
\multirow{3}{*}{M3} 
    & Trend     & 0.19 & 0.55 & 0.26 \\
    & Seasonal  & 0.18 & 0.37 & 0.45 \\
    & Residual  & 0.23 & 0.52 & 0.25 \\
\midrule
\multirow{3}{*}{M1} 
    & Trend     & 0.31 & 0.01 & 0.68 \\
    & Seasonal  & 0.44 & 0.25 & 0.31 \\
    & Residual  & 0.48 & 0.24 & 0.28 \\
\midrule
\multirow{3}{*}{M4} 
    & Trend     & 0.40 & 0.60 & 0.00 \\
    & Seasonal  & 0.40 & 0.50 & 0.10 \\
    & Residual  & 0.36 & 0.48 & 0.16 \\
\bottomrule
\end{tabular}
\end{table}

\subsubsection{Decomposition on M1 Monthly.}\label{sub_sec:m1_deco}

\begin{figure}[hbt]
    \centering
    \resizebox{\textwidth}{!}{\input{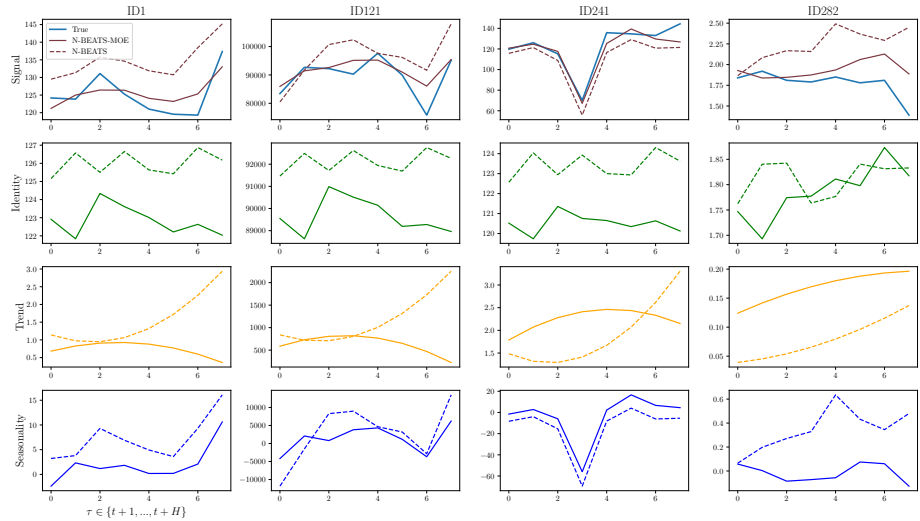}}
    \caption{Forecast decomposition in the M1 monthly dataset for series \textit{ID1}, \textit{ID121}, \textit{ID241}, and \textit{ID282} using the \texttt{N-BEATS-MOE} and \texttt{N-BEATS} models stack outputs. The gating values attributed by the \texttt{N-BEATS-MOE} model are approximately \([0.275, 0.169, 0.556]\), \([0.319, 0.251, 0.430]\), \([0.196, 0.193, 0.611]\), and \([0.366, 0.232, 0.402]\), respectively. Corresponding SMAPE values for \texttt{N-BEATS-MOE} are 2.68\%, 3.31\%, 4.40\%, and 10.15\%, while for \texttt{N-BEATS} they are 7.34\%, 8.03\%, 10.26\%, and 22.86\%. These results illustrate that a high gate softmax value does not necessarily imply greater output numeric scaling but indicates a greater contribution, in \%, to the output quality. 
}

    \label{fig:forecast_decomposition_m1}
\end{figure}

To assess the performance of \texttt{N-BEATS-MOE} on the M1 dataset, we analyze the decomposition of its three output stacks for the monthly frequency group and compare the resulting SMAPE scores with those of a comparable \texttt{N-BEATS} model.  We focus on series where our model achieved better performance (Figure~\ref{fig:forecast_decomposition_m1}). This comparison yields several insights. First, our method demonstrates a better understanding of the relative significance and appropriate scaling of each stack's output. For instance, in the time series \textit{ID1}, \texttt{N-BEATS-MOE} correctly identifies that the trend component is relatively unimportant, assigning it a weight of 16.93\% and scaling it down to the range of approximately $[0.4, 0.9]$. In contrast, \texttt{N-BEATS} overestimates this component’s importance, exaggerating its scale to around $[1, 3]$, likely due to overgeneralization. Second, our approach shows a superior ability to model seasonal fluctuations, including both sharp drops and spikes, as evidenced in series such as \textit{ID1}, \textit{ID121}, and \textit{ID241}. This capability appears to directly contribute to the improved forecast accuracy. Third, we observe that a high gate softmax value does not necessarily correspond to a high amplitude in the stack output, but rather reflects a significant qualitative contribution to the overall forecast accuracy.
These conclusions remain consistent across a broader set of series, including \textit{ID110}, \textit{ID112}, \textit{ID113}, \textit{ID115}, \textit{ID119}, \textit{ID120}, \textit{ID123}, \textit{ID242}, \textit{ID268}, and others.

\section{Conclusions}

In this paper, we present an extension of \texttt{N-BEATS}, a state-of-the-art deep learning approach for time series forecasting, by incorporating a MoE layer. More concretely, our architecture replaces the standard sum aggregation of the block outputs with a weighted sum, where the weights are determined by a gating network.

With this extension we add another layer of interpretability into the \texttt{N-BEATS}. By looking at the gate weights assigned by the gate it is possible to understand what stack the gate considers is the most important to the series. Furthermore, we hypothesize that this gating also helps the model handle heterogeneous datasets by allowing it to focus on different experts depending on the series characteristics.

We conducted experiments to evaluate the proposed approach on 12 benchmark datasets. The results show that the proposed approach can improve \texttt{N-BEATS}, especially in datasets composed of heterogeneous time series. In terms of alignment with the STL decomposition, the results where inconclusive. While in some cases the expert assignment aligned with the corresponding component, in other cases this did not happen. 

We believe that this approach offers a promising direction for building more interpretable and flexible forecasting models, especially in contexts where data exhibit diverse underlying patterns.




\end{document}